\documentclass{article}

\usepackage[final, nonatbib]{neurips_2021_ml4ps}

\usepackage[utf8]{inputenc} 
\usepackage[T1]{fontenc}    
\usepackage{hyperref}       
\usepackage{url}            
\usepackage{booktabs}       
\usepackage{nicefrac}       
\usepackage{microtype}      

\usepackage{amsmath}
\usepackage{amssymb}
\usepackage{amsfonts}
\usepackage{graphicx}
\usepackage[utf8]{inputenc}
\usepackage{xcolor}
\usepackage{caption}
\usepackage{subcaption}
\usepackage{wrapfig}

\usepackage{algorithm}
\usepackage{algpseudocode}

\newcommand{\beq}{\begin{equation}}
\newcommand{\eeq}{\end{equation}}
\newcommand{\bea}{\begin{eqnarray}}
\newcommand{\eea}{\end{eqnarray}}
\newcommand{\N}{\mathcal{N}}
\newcommand{\Neff}{\mathcal{N}_\textrm{eff}}
\newcommand{\dd}{\mathrm{d}}

\newcommand{\rob}[1]{{\color{purple} Roberto: #1}}

\title{Scaling Up Machine Learning For Quantum Field Theory with Equivariant Continuous Flows}

\author{
   Pim de Haan\thanks{Equal contribution}\\
   Qualcomm AI Research,\\ Qualcomm Technologies Netherlands B.V.\thanks{Qualcomm AI Research is an initiative of Qualcomm Technologies, Inc.}\\ University of Amsterdam
   \And
   Corrado Rainone$^*$\\
   Qualcomm AI Research,\\ Qualcomm Technologies Netherlands B.V.$^\dagger$\\
   \And
   Miranda C. N. Cheng\\
 Institute of Physics and \\
 Korteweg-de Vries Institute for Mathematics, \\University of Amsterdam, the Netherlands \\
 Institute for Mathematics, \\
 Academia Sinica, Taipei, Taiwan
   \And
   Roberto Bondesan\\
   Qualcomm AI Research,\\ Qualcomm Technologies Netherlands B.V.$^\dagger$
  }

\begin{document}

\maketitle


\begin{abstract}

We propose a continuous normalizing flow for sampling from the high-dimensional probability distributions of Quantum Field Theories in Physics. 
In contrast to the deep architectures used so far for this task, our proposal is based on a shallow design and incorporates the symmetries of the problem. 
We test our model on the $\phi^4$ theory, showing that it  systematically outperforms a realNVP baseline in sampling efficiency, with the difference between the two increasing for larger lattices.
On the largest lattice we consider, of size $32\times 32$, we improve a key metric, the effective sample size, from 1\% to 66\% w.r.t.~the realNVP baseline.

\end{abstract}
\begin{figure}[b!]
    \vspace{-1em}
     \centering
     \begin{subfigure}[t]{0.42\textwidth}
         \centering
         \includegraphics[width=\textwidth]{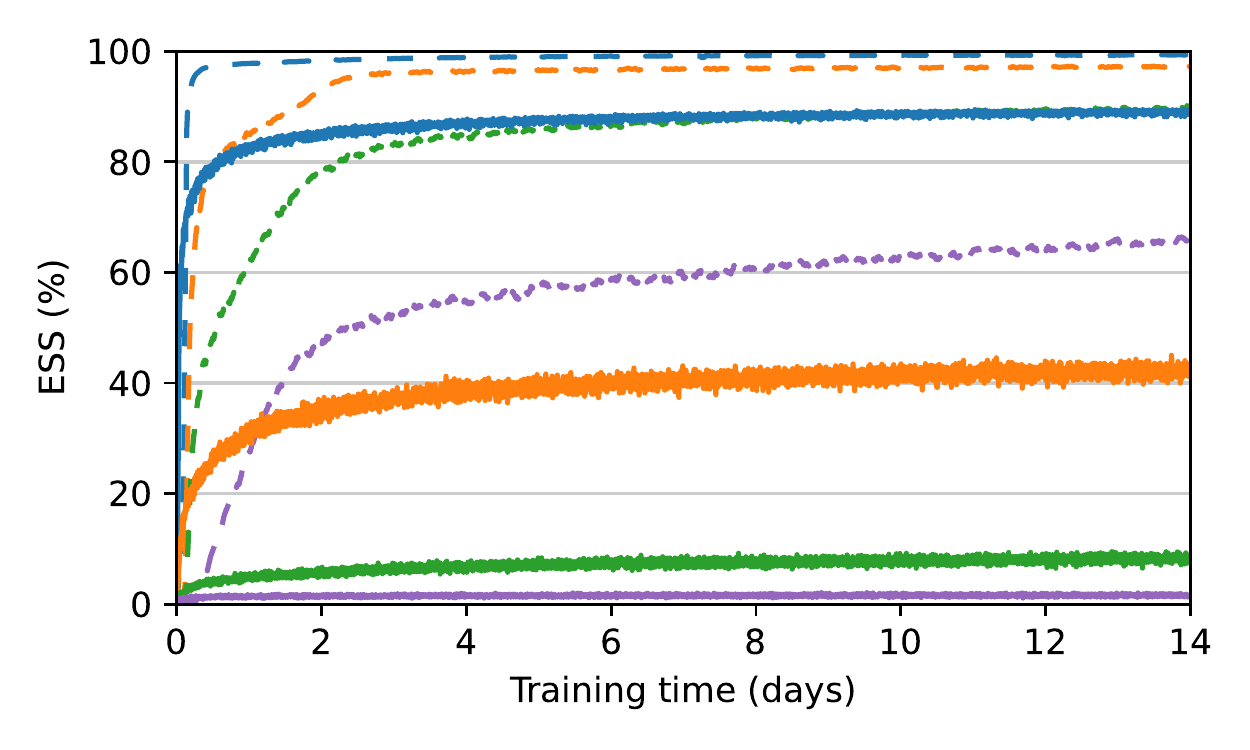}
         \label{fig:ess}
     \end{subfigure}
     \begin{subfigure}[t]{0.42\textwidth}
         \centering
         \includegraphics[width=\textwidth]{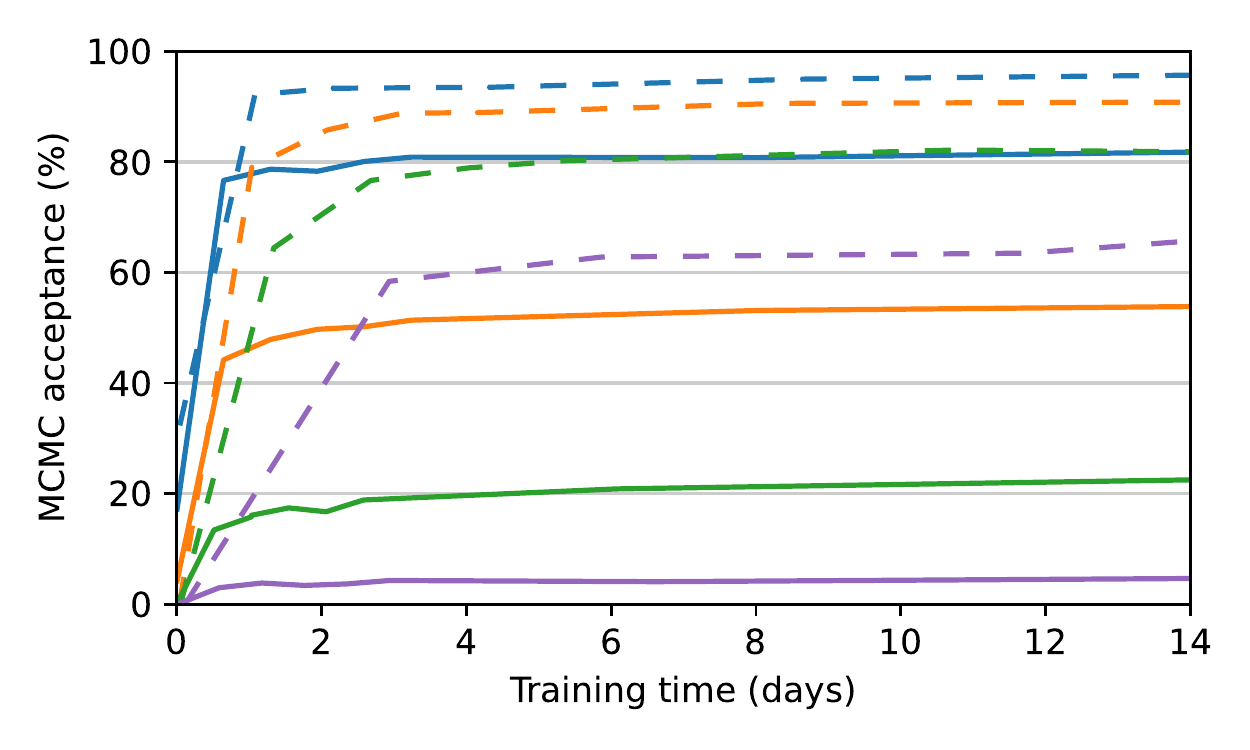}
         \label{fig:acceptance}
     \end{subfigure}
     \begin{subfigure}[t]{0.12\textwidth}
        \centering
        \includegraphics[width=\textwidth,trim={0 -3cm 0 0}]{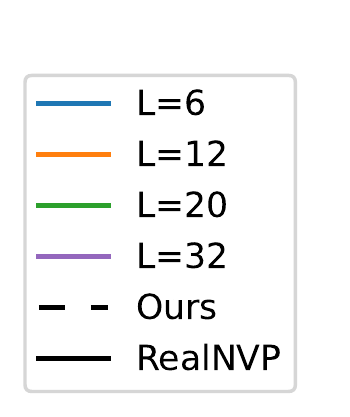}
     \end{subfigure}
     \caption{\small Effective Sample Size and MCMC acceptance rate for lattice size $L \times L$.}
     \label{fig:results}
\end{figure}

\vspace{-10pt}
\section{Introduction and Related Work}
\label{sec:intro}

Machine learning (ML) offers a novel tool which has the 
potential to outperform traditional computational methods in scientific applications, thanks to the ability of learning algorithms to improve automatically with more data and to adapt to the specific problem  at hand.
This has driven the fast adoption of ML in a variety of physical applications, ranging from quantum mechanics to molecular simulations  to particle physics. 
See for instance \cite{RevModPhys.91.045002,noe2020machine,radovic2018machine} for reviews.
In some cases, one can also show that ML methods are provably more efficient than traditional ones \cite{huang2021provably}.
Central research challenges in applying ML to physics include scaling up the ML models and interpretability. A general approach to these issues is to incorporate physical priors in the models, such as symmetries \cite{noe2020machine}.

In this paper we study the application of ML to lattice field theories. 
Quantum field theories can be used to describe diverse physical phenomena, ranging from properties of materials to fundamental forces of nature. 
Some of these theories are weakly coupled and can be studied via perturbative methods (i.e. Feynman Diagrams), but others are not, with as a notable example Quantum ChromoDynamics (QCD) which is an important pillar of the standard model of particle physics.
One of the major computational tools for these theories is to consider a lattice version of the theory, with the continuous space-time replaced by a discrete finite lattice. As is commonly done, we work in Euclidean space-time with no distinguished ``time" direction. 
In this setup, the main task is to sample from a known Boltzmann distribution, which is usually done via Markov Chain Monte Carlo (MCMC) methods~\cite{robert2013monte}.
 In all the applications, it is crucial that one should be able to extrapolate the lattice theories to the regime with a large number degrees of freedom, in order to access continuum physics or critical points. In many cases, however, the conventional MCMC methods suffer from the problem of Critical Slowing Down (CSD), namely that it takes a prohibitively long time to generate two independent distinct samples,
 which hampers robust extrapolations and renders the lattice field theory results unreliable.

Recently, a series of papers
(see for example \cite{albergo2019flow,Nicoli:2019gun,Pawlowski:2018qxs,PhysRevLett.125.121601,albergo2021flowbased,Nicoli:2020njz})
has started to explore ML techniques, such as Normalizing Flows, applied to such sampling tasks. The idea is that if we can learn an invertible map that trivializes an interacting model to a free theory, we can easily sample the latter and push back the samples through the inverse map to obtain (proposed) samples from the original non-trivial distribution. We can then correct for the mistakes of the learning algorithm by a Metropolis-Hastings accept-reject rule, as is typically done in MCMC.
In \cite{albergo2019flow,deldebbio2021efficient} the authors use a real NVP normalizing flow \cite{dinh2017density} as the invertible map for the $\phi^4$ theory distribution (defined below), and show that this technique can have better sampling efficiency than a pure MCMC method. Real NVP normalizing flows partition the lattice in two parts and at each layer modify only one part, which has the consequence that the symmetries of the square lattice are partially broken. 
Boltzmann generators \cite{noe2019boltzmann} extended these ideas to particle systems, further studied in
\cite{pmlr-v119-kohler20a} using symmetric continuous flows.
Refs.~\cite{PhysRevLett.125.121601, albergo2021flowbased} address the issue of incorporating gauge symmetries, and fermionic degrees of freedom, and
\cite{tomiya2021gauge} develops continuous flows for lattice QCD.

We summarize now the contributions of this paper:
\vspace{-7pt}
\begin{itemize}
    \item We extend \cite{pmlr-v119-kohler20a} and develop continuous normalizing flows for lattice field theories that are fully equivariant under lattice symmetries as well as the internal $\phi \mapsto -\phi$ symmetry of the $\phi^4$ model.
    Our flow is based on a shallow architecture which acts on handcrafted features.
    \item We train our model for the $\phi^4$ theory and 
    and for the $32\times 32$ lattice we improve the effective sample size from 1\% to 66\% w.r.t.~a real NVP baseline of similar size
    (Fig.~\ref{fig:results}).
    \item We study equivariance violations of real NVP models and contrast it with the exact equivariance of our flows.
\end{itemize}

\begin{figure}[b]
    \centering
    \includegraphics[width=0.9\textwidth,trim={1cm 0cm 3cm 1cm},clip]{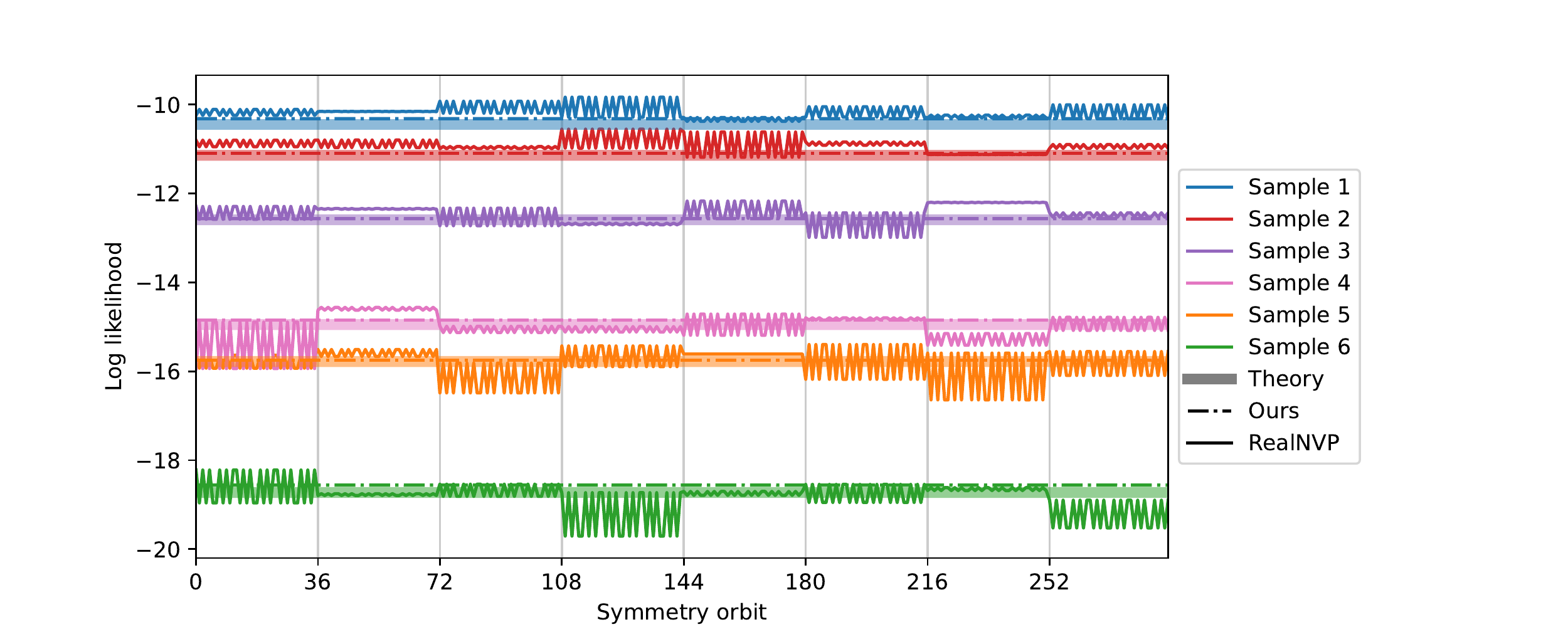}
    \caption{\small For 6 samples from an MCMC chain with $L=6$, we show 1) the true log likelihood given by the action, 2) the model log likelihood of RealNVP and 3) our model. The $x$-axis shows the likelihoods when the sample is transformed by all $8 * 6^2$ symmetries of the lattice. Within each block of 1/8th of the $x$-axis, the samples are related by a translation, and  by a rotation or mirror  between the blocks.}
    \label{fig:equivariance}
\end{figure}



\vspace{-0.5em}
\section{Background}
\vspace{-0.5em}
\label{sec:backgr}

\paragraph{Samplers Based on Normalizing Flows.}
The Metropolis-Hastings algorithm is a popular and flexible method to sample from a target density $p$ given a proposal distribution $q$ \cite{QuantumFields}. 
It proposes a sample $\phi'$ from $q$ and accepts it with probability $  \rho = \min(1,{q(\phi^{(i-1)})}{p(\phi')}/{p(\phi^{(i-1)})}{q(\phi')})$.
We report the detailed steps in algorithm~\ref{alg:MH}. CSD takes the form of long strings of rejections that lead to repeated samples in the chain, and therefore a reduction in sampling efficiency and quality of estimates. Conversely, when the proposal $q$ is identical to the target $p$, all proposals get accepted, and efficiency reaches its theoretical  maximum. 
The distribution
$q$ is usually painstakingly handcrafted with this goal in mind; the ML approach to this problem is conversely to \emph{learn} $q$, for example using a Normalizing Flow \cite{albergo2019flow}. In more detail, a normalizing flow is an invertible neural network $f$ that maps a  latent random variable $z$ distributed according to an easy-to-sample distribution $r(z)$ to a random variable $\phi$ whose distribution $q$ is the push-forward of $r$ under $f$  \cite{dinh2017density}.
To match $q$ with $p$, we can minimize the reverse KL divergence $\text{KL}(q|p)$: 
\begin{align}
    \text{KL}(q|p) 
    &=
    \mathbb{E}_{\phi \sim q} [\log \tfrac{q(\phi)}{p(\phi)}]
    =
    \mathbb{E}_{z \sim \rho} 
    [\log q(f^{-1}(z)) + S(f^{-1}(z))] + \log Z\,.
    \label{eq:rKL}
\end{align}
where we have used $-\log p(\phi) = S(\phi) + \log Z$. 
Since  $r$ is easy to sample, we can efficiently estimate Eq.~\eqref{eq:rKL} and minimize it w.r.t.~the parameters of $f$. 
As a result, a flow-based sampler first amortizes the cost of finding a good proposal distribution by training a normalizing flow $f$. We report the details of this training protocol in algorithm~\ref{alg:reverse_KL_training}.
The pushed forward 
distribution $q$ under $f$ is then used as proposal in Metropolis Hastings.

\begin{minipage}{0.4\textwidth}
\begin{algorithm}[H]
\caption{Independent Metropolis Hastings algorithm.}\label{alg:MH}
\begin{algorithmic}
\Require $q(\phi)$: proposal density, $p(\phi)$: target density.
\For{$i=1:K$}
    \State $\phi' \sim q(\phi)$
    \State $u \sim \text{Uniform}(u; 0,1)$
    \State $\rho = \min\left(1,\frac{q(\phi^{(i-1)})}{p(\phi^{(i-1)})}\frac{p(\phi')}{q(\phi')}\right)$
    \If{$u < \rho$}
        \State $\phi^{i} = \phi'$ \Comment{Accept}
    \Else
        \State $\phi^{i} = \phi^{i-1}$ \Comment{Reject}
    \EndIf
\EndFor
\end{algorithmic}
\end{algorithm}
\end{minipage}
\hfill
\begin{minipage}{0.5\textwidth}
\begin{algorithm}[H]
\caption{Reverse KL training of a normalizing flow}\label{alg:reverse_KL_training}
\begin{algorithmic}
\Require $f_\theta$: invertible neural network, 
$S(\phi)$: field theory action, $r(z)$ easy-to-sample distribution, $B$: batch size, $\gamma_i$: learning rates.
\For{$i=1:K$}
    \State $z^1,\dots,z^B \sim r(z)$
    \State $L =
    \tfrac{1}{B}\sum_{b=1}^B
    \log q(f_\theta^{-1}(z^b)) + S(f_\theta^{-1}(z^b))$\\
    \State $\theta = \theta - \gamma_i \nabla_\theta L$
\EndFor
\end{algorithmic}
\end{algorithm}
\end{minipage}

In \cite{albergo2019flow} it is shown that such a sampler achieves competitive results on small lattices for the $\phi^4$ theory. Specifically, a Real NVP flow uses a stack of coupling layers $g : z\in \mathbb{R}^N \mapsto \phi \in \mathbb{R}^N$ defined as follows. We partition the input/output space in two parts of size $N_a, N_b$, and 
use indices $a$ and $b$ for the elements in each part. Then a coupling layer is \cite{dinh2017density}:
\begin{align}
    \phi_a = z_a\,,\quad
    \phi_b = z_b \odot s(z_a) + t(z_a)\,.
\end{align}
Here $s, t$ are neural networks, and 
this layer is invertible if $s$ is invertible for all $t$.
Its inverse, $g^{-1}$, is 
\begin{align}
    z_a = \phi_a \,,\quad
    z_b = (\phi_b - t(\phi_a)) \odot s(\phi_a)^{-1}\,.
\end{align}
Denoted $f$ the real NVP normalizing flow with latent distribution $\rho$, 
we can compute explicitly the push forward distribution entering the reverse KL loss:
\begin{align}
    \log q(f^{-1}(z)) 
    =
    \log \rho(z) 
    -
    \log \left|\det J_f \right|  
\end{align}
where the log Jacobian determinant can be computed in $O(N)$ time instead of a naive $O(N^3)$, by summing each coupling layer  contribution:
\begin{align}
    \log \left|\det J_f(z) \right|
    =
    \sum_\ell 
    \sum_{i=1}^{N_a} \log |s_\ell(z_a)_i|
    \,,
\end{align}
where $s_\ell$ is the scale factor of the $\ell$ coupling layer.
While models based on real NVP flows have achieved success in a number of ML tasks, see e.g.~\cite{glow}, we note that partitioning the input space prevents to construct an equivariant layer under a symmetry that maps $(z_a)_i \mapsto (z_b)_j$ for some $i,j$. 


In a continuous normalizing flow, we define an invertible mapping $f :  \mathbb{R}^N \to \mathbb{R}^N$, $z \mapsto \phi$, as the flow under a neural ODE~\cite{chen2018neural} for a fixed time $T$:
\beq
\frac{\dd x(t)}{\dd t} = g(x(t),t;\theta)~{\rm with}~x(t)\lvert_{t=0}= z,~ x(t)\lvert_{t=T}=\phi. 
\label{eq:ODE}
\eeq
Here the vector field $g(x(t),t;\theta)$ is a neural network with weights $\theta$.
The log probability then follows another ODE~\cite{chen2018neural}
\beq
\frac{\dd\log p(x(t))}{\dd t} = -(\nabla_x \cdot g)(x(t),t;\theta)~
{\rm with} ~ p(x(t))\lvert_{t=0}= r(z),~ p(x(t))\lvert_{t=T}= q(\phi) .
\label{eq:logpODE}
\eeq
$g$ is not constrained as in the real NVP case, which allows one to build in symmetries more easily. Following \cite{pmlr-v119-kohler20a}, if the the vector field $g$ is equivariant, the resulting distribution on $\phi$ is invariant. We will show in the next section how to construct a $g$ equivariant to the square lattice symmetries.

\paragraph{The $\phi^4$ theory.}
Our goal is to improve the sampling performance for the so-called $\phi^4$ theory. It is a relatively simple quantum field theory which nevertheless possesses non-trivial symmetry properties and an interesting phase transition, which make it an ideal testing ground for new computational ideas. 
In the case of $\phi^4$ theory in two dimensions, the {\em field configuration} is a real function on the vertex set $V_L$ of the square lattice with periodic boundaries and size $L\times L$: $\phi: V_L \to {\mathbb R}$. 
The $\phi^4$ theory is described by a probability density $p(\phi) = \exp(-S(\phi))/Z$, with action 
\beq
S(\phi) = \sum_{x,y\in V_L}\phi(x) \Delta_{x,y} \phi(y) + \sum_{x\in V_L}m^2\phi(x)^2 + \mathit{\lambda}\phi(x)^4
\label{eq:S}
\eeq
In the above,
$\Delta$ is Laplacian matrix of the square lattice $({\mathbb Z}/L{\mathbb Z})^{\times 2}$, $m$ and $\lambda$ are numerical parameters. In the case of this and other non-trivial field theoretical densities, direct sampling is impossible due to the statistical correlation between  degrees of freedom spatially separated up to the \emph{correlation length} of the theory, a fundamental quantity denoted as $\xi$;
$Z$ is the normalisation factor that is not known analytically for $\lambda\neq 0$.   
Note that, besides the (space-time) symmetries of the periodic lattice, the theory possesses a discrete global symmetry $\phi \mapsto -\phi$. We shall choose the couplings in such a way that only one minimum of the action, invariant under this symmetry, exists. See \cite{hackett2021flowbased} for relevant work in the case of a symmetry-broken case. 

\section{Method\label{sec:arch}}
\begin{figure}[h!]
    \centering
    \includegraphics[width=0.5\textwidth]{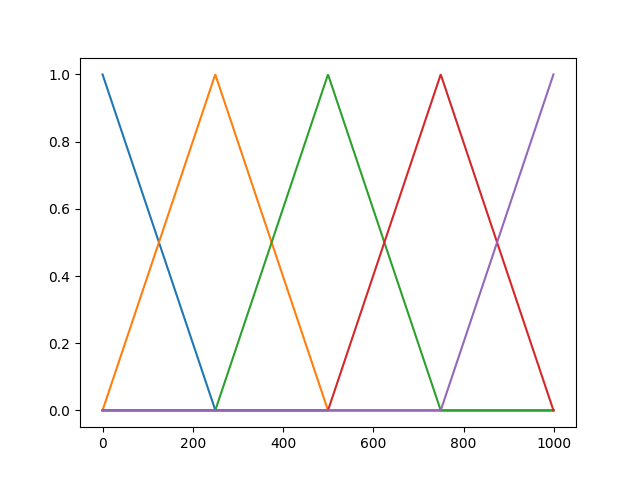}
    \caption{The 5D linear interpolation kernel.}
    \label{fig:linear-interpolation-kernel}
\end{figure}

\begin{figure}[b!]
    \centering
    \includegraphics[width=0.5\textwidth]{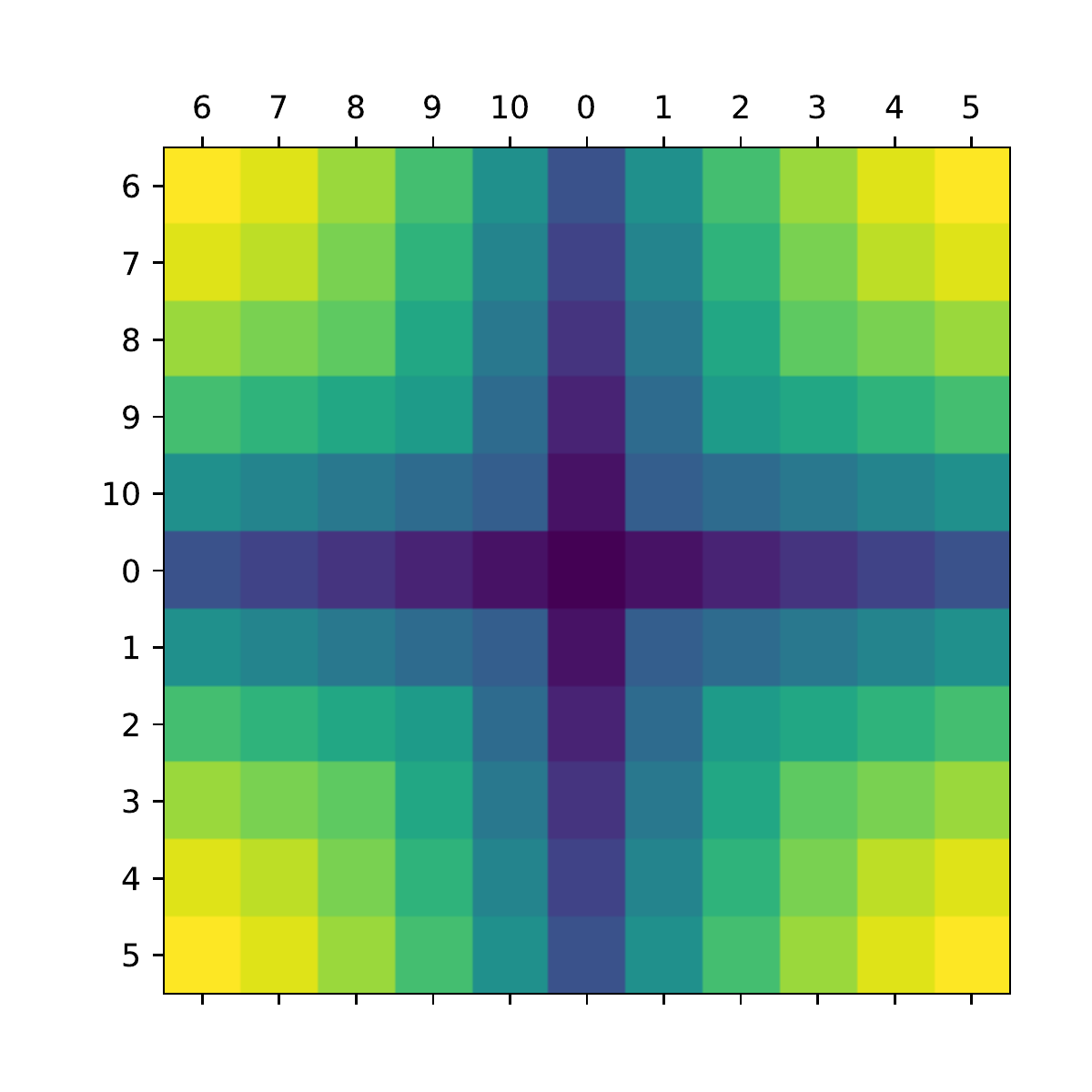}
    \caption{Each color denotes one $D_4$ orbit of a periodic $L=11$ lattice, leaving $x_0=(0, 0)$ invariant. For each color in the figure, and for each dimensions $a$ and $f$, we have a free parameter $W_{[y]af}$.}
    \label{fig:orbits}
\end{figure}

Inspired by equivariant flows used for molecular modelling \cite{pmlr-v119-kohler20a}, we propose to use the following vector field for the neural ODE:
\begin{align}
\label{eq:our_model}
\frac{\dd \phi(x, t)}{\dd t} = \sum_{yaf} W_{xyaf} K(t)_a \sin(\omega_f \phi(y, t))
\end{align}
and we have some learnable frequencies $\omega_f$, initialised standard normal, to construct a Fourier basis expansion inspired by Fourier Features~\cite{tancik2020fourier}.
We chose only sines to enforce $\phi\mapsto -\phi$ equivariance. 
In Eq.~\eqref{eq:our_model} $x,y\in V_L$, $K(t)_a$ is a linear interpolation kernel, illustrated in Fig.~\ref{fig:linear-interpolation-kernel} for the 5-dimensional case. $W$ is a learnable tensor, initialised to $0$, so that the initial flow in an identity transformation. The divergence of this simple flow is computed analytically.
The periodic lattice $V_L$ has spatial symmetry group $G=C_L^2 \rtimes D_4$, the semi-direct product of two cyclic groups $C_L$ of translations and Dihedral group $D_4$ of right angle rotations and mirrors. To ensure spatial equivariance of the vector field model, we should have that $\forall g \in G, x,y,a, f, W_{g(x)g(y)af}=W_{xyaf}$. 
Using the translation subgroup, we can map any point $x$ to a fixed point $x_0$. This allows us to write $W_{xyaf}=W_{x_0t_x(y)af}$, $t_x(y) = y - x + x_0$.
Then let $H\simeq D_4$ be the subgroup of $G$ such that $g(x_0)=x_0$ for all $g\in G$, and denote the orbit of $y$ under $H$ by $[y]=\{y' \mid \exists g \in H, g(y)=y'\}$. For each such orbit $[y]$ and dimension $a$ and $f$, a free parameter $W_{[y]af}$ exists, so that the other parameters are generated by $W_{xyaf}=W_{[t_x(y)]af}$.
As most orbits are of size 8, the number of free parameters per $a$ and $f$ is approximately $L^2/8$.
See Fig.~\ref{fig:orbits} for a figure of the orbits.

\vspace{-0.5em}
\section{Experiments\label{sec:exps}}
\vspace{-0.5em}
We here discuss experimental results obtained using the flow architecture just described for the $\phi^4$ theory. The flow architecture we propose is trained according to the following, and fairly standard experimental protocol. Starting field configurations $z$, composed of $D\equiv L\times L$ uncorrelated sites are sampled from a Gaussian prior of zero mean and unit variance. They are then passed through the flow, which integrates its ODE from $x(t=0) \equiv z $ to the transformed field configurations $x(t=1) \equiv \phi$, as well as the ODE for the associated log Jacobian. We opted to use the 4th order Runge-Kutta solver, with a relatively generous number (50) of integration steps thanks to the simplicity of our flow. The batch size was fixed at $M=100$ throughout. The parameters of the flow are trained via minimization of the reverse KL loss, eq.~\eqref{eq:rKL}, which we perform with the Adam~\cite{kingma2014adam} optimizer with a learning rate fixed to $\eta=0.001$ for the first 250 steps, and reduced by $\frac{1}{10}$ afterwards.

For all lattice sizes, we used in our model the same hyperparameters: 10 time dimensional time kernel $K$ and $9$ learnable frequencies $w_f$. We experimented also with 3 time dimensions and found worse performance, but didn't run any further hyperparameter search. During some exploratory experiments, we found indications that using larger batch sizes, like 1000 or larger, and/or other learning rate scheduling may speed up training by a significant amount. The RealNVP model uses the code and hyperparameters from \cite{albergo2021introduction} (CC BY 4.0 licensed), which uses 16 coupling layers and in each coupling layer a 2 layer CNN with 8 dimensional hidden layers and kernel size 3. All computations were run on NVidia V100 GPUs. The training for the results reported in \ref{tab:params} lasted two weeks. The expected sample sizes were computed over 1000 samples. The ESS plots were smoothed over by a rolling average over 200 epochs to make trends more readable. Each epoch lasted 50 training steps, each of which had batch size 100.

In order to investigate scalability, we consider a range of lattice side lengths $L$ going from 6 to 32 with the correlation length being indirectly fixed at $L/4$ via appropriate choice of the coupling constant $\lambda$ in the action, Eq.~\eqref{eq:S} \footnote{Here and everywhere else we work with unit lattice spacing. }. The choice of $m,\lambda$ for all sizes are reported in table~\ref{tab:params}.
\begin{wraptable}{r}{0.55\textwidth}
    \centering
    \begin{tabular}{c|c c c c}\hline\hline
         $L$ & 6 & 12 & 20 & 32 \\ \hline
         $m^2$ & -4 & -4 & -4 & -4\\
         $\lambda$ & 6.975 & 5.276 & 5.113 & 4.750 \\
        $\frac{L}{\xi}$ & 3.98 & 3.99 & 4.05 & 4.05 \\
        $\chi$ & 1.06 & 4.12 & 10.57 & 25.34 \\
        Acceptance (\%) & 96 & 91 & 82 & 66 \\
        ESS (\%) & 99 & 97 & 90 & 66 \\
         \hline
    \end{tabular}
    \caption{\small Observables computed with our model.}
    \vspace{-1em}
    \label{tab:params}
\end{wraptable}
To assess model quality, we use two different but related metrics, the Effective Sample Size (ESS)
\begin{equation}
\label{def:ESS}
\textrm{ESS} \equiv \frac{ \left(\frac{1}{\N} \sum_i p(\phi_i)/q(\phi_i) \right)^2 }{ \frac{1}{\N} \sum_i \left( p(\phi_i)/q(\phi_i) \right)^2 },
\end{equation}
and the MCMC acceptance rate. A perfectly trained flow would be able to produce a set of uncorrelated field configurations $(\phi_i)_{i=1}^\N$, which could then be used to compute estimates of physical observables whose MSE would scale like $1/\sqrt{\N}$. This is usually not the case in practice, and the MSE actually scales like $1/\sqrt{\Neff}$, with $\Neff<\N$; the ratio $\frac{\Neff}{\N}$ is precisely the ESS, which is the metric we use to monitor the progress of training. At test time, conversely, we evaluate the trained flows by using them to generate a chain of MCMC proposals of length $\N= 10^6$, and use the fraction of accepted moves as our test metric.

In the left panel of Figure~\ref{fig:results}  we report the ESS values against the real training time (in days) with the realNVP baseline for comparison, which is based on \cite{albergo2021introduction}.
Our model attains much larger ESSs and acceptance rates in much shorter times, for all sizes $L$, even taking into account the fact that each of its training steps involves integration of an ODE. In the right panel we record the acceptance rates of our model, which are similarly superior across different sizes, indicating that their performance suffers  less from CSD. 
We also cross-check the quality of the generated samples via the estimation of two physical observables: 
the \emph{two point susceptibility} $\chi_2$
\begin{equation}
\chi_2 \equiv \frac{1}{D}\sum_{x,y} \mathbb{E}\left[\phi(x)\phi(x+y)\right] - \mathbb{E}\left[\phi(x)\right]\mathbb{E}\left[\phi(x+y)\right]. 
\label{eq:obs}
\end{equation}
and the \emph{inverse pole mass} $m_p L = \frac{L}{\xi}$. Explicitly, we use the following estimator for $\chi_2$:
\begin{equation}
\begin{split}
\hat{G}(x) \equiv\ \frac{1}{D}\sum_{y}&\frac{1}{\N}\sum_{i=1}^\N\left[\phi_i(y)\phi_i(y+x) - \phi_i(x+y)\frac{1}{\N}\sum_{j=1}^\N \phi_j(x)\right.\\
&\left. - \phi_i(x)\frac{1}{\N}\sum_{j=1}^\N \phi_j(x+y) + \frac{1}{\N}\sum_{k=1}^\N \phi_k(x)\frac{1}{\N}\sum_{j=1}^\N \phi_j(x+y)\right];
\end{split}
\end{equation}
and the susceptibility is then simply obtained as $\hat{\chi}_2 = \sum_{x} \hat{G}(x)$. As for the inverse pole mass, the estimator we use is:
\begin{equation}
\hat{m}_p \equiv \frac{1}{L-1}\sum_{x_2=1}^{L-1}\left(\frac{G_c(x_2-1)+G_c(x_2+1)}{2G_c(x_2)} \right),
\end{equation}
where
\begin{equation}
G_c(x_2) \equiv \frac{1}{L}\sum_{x_1=1}^{L}\hat{G}(x_1,x_2),
\end{equation}
and the vertices of the periodic lattice are explicitly labelled by their two coordinates $(x_1,x_2)$.
Given the choice of parameters detailed above \cite{vierhaus}, we expect $m_p L\simeq 4$ for every $L$, which is always the case. 
The resulting values of our computations are recorded in Table \ref{tab:params}.

Furthermore, we verify the invariance of the model distribution $q(\phi)$, which is a result of the equivariance of our model, under the spatial symmetries.  In figure~\ref{fig:equivariance} we report the model log-probability of some randomly chosen field configurations and their images under spatial symmetry transformations. As expected, these symmetry-equivalent configurations are equally likely, while  this is not the case for the realNVP. 
Interestingly, the symmetry  violations of 
the latter does not seem to decrease as training progresses, indicating that these exact symmetries are not easily learned.

\begin{figure}[t]
    \centering
    \includegraphics[width=\textwidth]{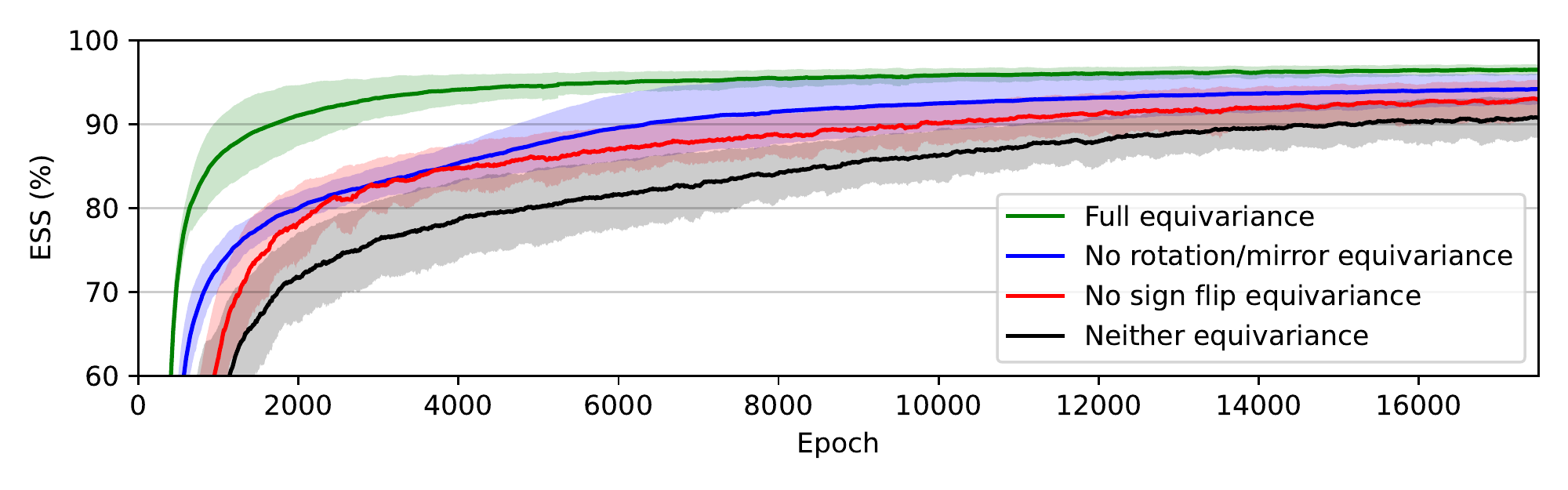}
    \caption{Ablation study on the $L=12$ lattice, varying the equivariance properties. Shown are mean and standard deviation across three training runs per variation.}
    \label{fig:ablation}
\end{figure}

Finally, we conducted an ablation study to verify that equivariance is beneficial. We consider two modifications to the model. In one variation, we remove the rotational and mirror spatial symmetries, by constraining the $W$ tensor in Eq.~\eqref{eq:our_model} to be equivariant only to translations, but not to rotations and mirrors. 
In the second variation, we remove the sign flip equivariance by replacing the sine in Eq.~\eqref{eq:our_model} by a combination of sine, cosine and a constant. The results in figure~\ref{fig:ablation} show that removing either or both of the equivariance properties harms performance, indicating that incorporating all symmetries in the model is indeed beneficial.

\vspace{-0.5em}
\section{Conclusion\label{sec:concl}}
\vspace{-0.5em}
In this paper, a shallow normalizing flow with features handcrafted via domain-specific knowledge, few trainable parameters, and an equivariant structure, is shown to significantly outperform a deep architecture trained end-to-end on a task of generative modelling of a simple, but paradigmatic field theory. 
There are various worthwhile directions for further investigations.  We now discuss two of them. The first is to systematically assess the scalability of our approach, as is done in~\cite{deldebbio2021efficient} for a seemingly less scalable architecture. The scalability will tell whether our method actually has the potential to provide a pragmatic solution to CSD. 
The second  is to include equivariance under \emph{local} and \emph{non-abelian} symmetries to our framework, which is a necessary step towards an ML-based sampler for the more challenging and interesting case of QCD. One might believe that the simplicity and flexibility of our proposal suggests promising future results. 


\section*{Acknowledgements}
We would like to thank Vassilis Anagiannis for his contributions to the early stages of this project and Jonas K\"ohler and Max Welling for their helpful discussions.

\bibliographystyle{unsrt}
\bibliography{refs}
\newpage

\end{document}